\documentclass{article}

\usepackage{microtype}
\usepackage{graphicx}
\usepackage{subcaption}
\usepackage{booktabs} % for professional tables
\usepackage{bm}
\usepackage{hyperref}
\usepackage{booktabs}
\usepackage{xcolor}
\newcommand{\best}[1]{\textbf{\textcolor{red}{#1}}}
\newcommand{\secondbest}[1]{\underline{\textcolor{blue}{#1}}}

\usepackage[preprint]{temp}
\renewcommand{\today}{\mbox{}}

\usepackage{amsmath}
\usepackage{amssymb}
\usepackage{mathtools}
\usepackage{amsthm}

\usepackage[capitalize,noabbrev]{cleveref}

\theoremstyle{plain}

\theoremstyle{definition}

\theoremstyle{remark}

\usepackage[textsize=tiny]{todonotes}

\begin{document}
	
	\twocolumn[
	\title{LightGTS-Cov: Covariate-Enhanced Time Series Forecasting}

	\begin{tempauthorlist}
		\tempauthor{Yong Shang}{1}
		\tempauthor{Zhipeng Yao}{1}
		\tempauthor{Ning Jin}{1}
		\tempauthor{Xiangfei Qiu}{2}
		\tempauthor{Hui Zhang}{1}
		\tempauthor{Bin Yang}{2}
	\end{tempauthorlist}
	
	\affiliation{1}{Inspur Group Co. Ltd, Shandong Inspur Database Technology Co., Ltd, Shandong, China}
	\affiliation{2}{School of Data Science and Engineering, East China Normal University, Shanghai, China}
	
	\correspondingauthor{Ning Jin}{jinning@inspur.com}

	\vskip 0.3in
	]
	\printAffiliationsAndNotice{}

	\begin{abstract}
		
		Time series foundation models are typically pre-trained on large, multi-source datasets; however, they often ignore exogenous covariates or incorporate them via simple concatenation with the target series, which limits their effectiveness in covariate-rich applications such as electricity price forecasting and renewable energy forecasting.
		We introduce \textbf{LightGTS-Cov}, a covariate-enhanced extension of LightGTS that preserves its lightweight, period-aware backbone while explicitly incorporating both past and future-known covariates.
		Built on a $\sim$1M-parameter LightGTS backbone, LightGTS-Cov adds only a $\sim$0.1M-parameter MLP plug-in that integrates time-aligned covariates into the target forecasts by residually refining the outputs of the decoding process.
		Across covariate-aware benchmarks on electricity price and energy generation datasets, LightGTS-Cov consistently outperforms LightGTS and achieves superior performance over other covariate-aware baselines under both settings, regardless of whether future-known covariates are provided. 
		We further demonstrate its practical value in two real-world energy case applications: long-term photovoltaic power forecasting with future weather forecasts and day-ahead electricity price forecasting with weather and dispatch-plan covariates. Across both applications, LightGTS-Cov achieves strong forecasting accuracy and stable operational performance after deployment, validating its effectiveness in real-world industrial settings.

	\end{abstract}

	\section{Introduction}
	Time series forecasting is a core capability in many high-impact domains~\cite{hittawe2024time,sezer2020financial,nowotarski2018recent,yin2016forecasting,morid2023time}, where the quality, completeness, and usability of data largely determine achievable performance. In industrial systems, forecasts rarely rely on target history alone; instead, they are driven by a combination of historical target observations and exogenous covariates, such as meteorological variables, calendar effects, and dispatch or schedule information~\cite{hyndman2018forecasting,maccaira2018time}. Moreover, many real-world applications provide future-known covariates---e.g., weather forecasts and pre-announced operating plans---that offer horizon-specific signals and become increasingly critical as the forecast horizon extends. Ignoring these covariates can substantially degrade long-term stability and peak accuracy. This issue is further amplified under regime changes, where purely autoregressive rollouts tend to accumulate errors and drift~\cite{taieb2012review}.
	
	These considerations are particularly prominent in energy systems, a covariate-rich setting where forecasts directly support operational decisions~\cite{mystakidis2024energy,christen2020exogenous}. In this paper, we focus on two representative industrial scenarios.
	First, for photovoltaic (PV) power forecasting~\cite{akhter2019review,gupta2021pv} at a real plant, the objective is to produce multi-day generation trajectories using recent output history, on-site environmental measurements, and future-known weather forecasts. PV generation follows strong diurnal and seasonal regularities but is highly sensitive to irradiance and cloud evolution, making long-term forecasting prone to drift and peak bias if future weather information is not effectively utilized. 
	Second, for electricity price forecasting~\cite{weron2014electricity,lago2021forecasting} in market operations, the goal is to predict next-day price curves using historical prices alongside exogenous drivers such as weather and dispatch disclosures. Prices exhibit a pronounced intraday peak--valley structure yet remain highly volatile with occasional regime shifts. In particular, time-critical releases of next-day dispatch plans provide horizon-specific signals that reflect expected supply--demand balance under market clearing; failing to leverage such information---or fusing it naively---often yields inaccurate peak/trough dynamics and unstable forecasts.
	
	Recent advances in time series foundation models (TSFMs)~\cite{timesfm,Chronos,moment} have demonstrated strong cross-domain generalization through large-scale pretraining. However, deploying TSFMs in covariate-rich industrial scenarios still faces practical gaps. Many state-of-the-art TSFMs are large (often tens to hundreds of millions of parameters), making fine-tuning, inference cost, and system integration non-trivial in production settings. In addition, many pretrained TSFMs adopt target-centric or channel-independent formulations~\cite{han2024capacity} and provide limited native support for structured integration of historical and future-known covariates. As a result, covariates are often added via simple concatenation or task-specific engineering, which can degrade peak accuracy and long-term stability in production.
	
	Existing covariate-aware adaptation methods are typically co-designed with a specific TSFM backbone and covariate schema, and their robustness can vary when transferred across architectures or information sets~\cite{qin2025cora,han2025unica,arango2025chronosx,benechehab2025adapts,yamaguchi2025citras}. 
	Moreover, a one-size-fits-all plug-in is difficult to realize in practice, as it must accommodate heterogeneous covariates and diverse backbone designs without sacrificing simplicity and deployment stability.
	Motivated by these observations, we prioritize an industry-ready covariate integration that is lightweight, stable, and controllable.
	Accordingly, we build on LightGTS~\cite{lightgts}---a compact (\(\sim\)1M-parameter) backbone that learns strong periodic structure via period-aware tokenization and parallel, non-autoregressive decoding---to enable low-latency inference and reliable adaptation in production environments.
	Our design is also architecture-friendly: it operates on decoder-side, time-aligned latent tokens and introduces covariate information through lightweight MLP-based fusion, avoiding intrusive changes to attention mechanisms or training objectives and making the module easy to port to other backbones that expose decoder representations.
	
	In this paper, we propose \textbf{LightGTS-Cov}, a covariate-enhanced extension of LightGTS that preserves simplicity and deployment efficiency while substantially improving its ability to leverage exogenous signals in covariate-rich energy forecasting.
	LightGTS-Cov processes past covariates through the same period-aware tokenization as the target to obtain aligned decoder-side representations, and embeds future-known covariates into horizon-aligned patch embeddings.
	On top of the decoded target tokens, two lightweight post-decoder MLP blocks perform a two-stage fusion: the first performs past-covariate fusion with decoded target tokens, and the second performs future-covariate fusion using the horizon-aligned embeddings to produce covariate-refined tokens. Finally, the decoded target tokens and the covariate-refined tokens are projected by the shared LightGTS output head and summed to produce the final forecast.

	Our main contributions are summarized as follows:
	\begin{itemize}
		\item We propose a practical covariate-enhanced extension of LightGTS that supports both historical and future-known covariates, and provides a unified, time-aligned pathway to integrate exogenous signals into forecasting.
		\item We introduce a lightweight post-decoder MLP fusion design (\(\sim\)0.1M parameters) that refines decoder-side target tokens via two-stage covariate fusion and aggregates the result through a shared output head, without redesigning the backbone or altering its decoding procedure.
		\item Extensive experiments show that LightGTS-Cov consistently improves over LightGTS and remains competitive with strong covariate-aware baselines on electricity price and energy generation benchmarks, and further demonstrates stable effectiveness in two real-world industrial deployments.
	\end{itemize}

	\section{Related Work}
	
	\subsection{Time Series Foundation Models}
	% Recent years have seen rapid progress in time series foundation models (TSFMs), which learn transferable temporal representations through large-scale pretraining and can generalize to downstream tasks in a zero-shot or low-shot manner. One line of work adapts large language models (LLMs) for forecasting by reprogramming time series into prompt- or token-like inputs while keeping most backbone parameters frozen~\cite{gruver2023large,cao2023tempo,xue2023promptcast}, as exemplified by Time-LLM~\cite{jin2023time}.
	Recent years have seen rapid progress in time series foundation models (TSFMs), which are pretrained from scratch on large time-series corpora with architectures tailored to temporal data. For instance, TimesFM~\cite{timesfm} and Timer~\cite{liu2024timer} adopt decoder-only Transformer formulations with next-token prediction objectives, while Chronos~\cite{Chronos} discretizes real-valued series via scaling and quantization and performs language-model-style prediction in the token space.
	Sundial~\cite{liu2025sundial} further targets probabilistic forecasting by incorporating generative modeling to enhance distributional flexibility. 
	Beyond forecasting-centric objectives, MOMENT~\cite{moment} investigates masked time-series modeling as a general pretraining paradigm for time-series analysis and downstream transfer. Recent work has also explored lightweight general forecasters. 
	LightGTS~\cite{lightgts} is a compact ($\sim$1M-parameter) general time-series forecasting model that captures periodic structure via period-aware tokenization and parallel decoding, and serves as the backbone enhanced in this paper. 
	In addition, ROSE~\cite{wang2024rose} improves general forecasting through register-assisted modeling and decomposed frequency learning, demonstrating the benefit of explicitly structuring frequency components for transferable time-series representations. Many TSFMs are pretrained in a univariate or channel-independent manner due to the heterogeneity of large-scale corpora, which often limits native support for exogenous covariates in downstream forecasting~\cite{timesfm,Chronos,moment}.
	A natural direction is therefore to extend TSFMs to natively handle multivariate inputs and covariate-informed contexts, as explored by models such as Moirai~\cite{Moirai}, Chronos-2~\cite{chronos2}, and Timer-XL~\cite{timerxl}. 
	% Moirai is among the few TSFMs that explicitly model cross-variable interactions, Chronos-2 broadens the forecasting scope with multivariate and covariate-aware forecasting, and Timer-XL further extends decoder-only Transformers to a unified forecasting framework via TimeAttention, supporting univariate, multivariate, and covariate-informed forecasting.

	\subsection{Covariate-aware Deep Forecasting Models}
	% {\color{blue}In real-world forecasting, exogenous drivers often provide predictive signals beyond the target history; consequently, covariates have long been incorporated to improve accuracy and robustness.
	% Classical statistical approaches such as ARIMAX and SARIMAX treat covariates as exogenous regressors within a structured forecasting framework.}
	Deep learning forecasters provide expressive mechanisms for leveraging heterogeneous covariates. For example, CrossLinear~\cite{zhou2025crosslinear} proposes a cross-correlation embedding to inject exogenous variables into endogenous representations with low overhead. TFT~\cite{lim2021temporal} performs structured fusion of static features, observed covariates, and known future inputs using dedicated encoders together with gating and attention mechanisms.
	Models such as NBEATSx~\cite{NBEATSx} and TiDE~\cite{das2023long} explicitly exploit future-known covariates to improve multi-step forecasting, while TimeXer~\cite{wang2024timexer} strengthens target-covariate interactions in Transformer-style architectures by modeling the target at the patch level and incorporating covariates via sequence-level representations.

	\subsection{Adaptation Methods for TSFMs}
	For TSFMs, covariate integration is challenging because downstream applications exhibit diverse covariate schemas (e.g., variable counts, modalities, sampling rates, and availability as past-only or future-known). A complementary line of work therefore focuses on parameter-efficient adaptation~\cite{pfeiffer2023modular,hu2022lora,houlsby2019parameter}, augmenting pretrained backbones with lightweight modules to incorporate covariates without full retraining.
	Representative examples include ChronosX~\cite{arango2025chronosx}, which introduces modular injection blocks for past observed and future-known covariates, and CoRA~\cite{qin2025cora}, which keeps the backbone largely frozen and injects covariate information through minimally intrusive modules guided by estimated covariate relevance.
	UniCA~\cite{han2025unica} further targets general covariate-aware forecasting by standardizing heterogeneous covariates into a unified representation before integration, while AdaPTS~\cite{benechehab2025adapts} adapts pretrained univariate backbones toward multivariate (and probabilistic) forecasting through parameter-efficient transformations.
	
	Such plug-in style adaptations are attractive in practice because they incorporate covariates with limited additional parameters and reduced disruption to pretrained representations.
	In contrast to backbone-coupled or structurally invasive injections, our LightGTS-Cov adopts a non-intrusive, decoder-side residual plug-in: it introduces covariate information only through time-aligned latent tokens after the LightGTS decoding process, leaving the original period-aware patching and parallel decoding pipeline unchanged.
	Moreover, by explicitly aligning both past covariates and future-known covariates at the token level, the plug-in provides a unified and controllable interface for heterogeneous exogenous inputs with minimal overhead ($\sim$0.1M parameters on top of a $\sim$1M backbone), making it particularly suitable for production settings where stability and low-latency inference are critical.

	\section{Methodology}
	
	\subsection{Problem Settings}
	Let $\bm{X}^{target} \in \mathbb{R}^{C \times T}$ denote the target time series with $C$ variables and a historical context length $T$.
	Let $\bm{X}^{past} \in \mathbb{R}^{M_p \times T}$ denote the associated past covariates observed over the same history window (e.g., on-site sensors).
	Over a forecast horizon of length $F$, we further assume access to future-known covariates $\bm{Y}^{future} \in \mathbb{R}^{M_f \times F}$ (e.g., weather forecasts, pre-announced dispatch plans). 
	For covariates available both historically and in the future (e.g., weather), their historical segment is included in the past covariates $\bm{X}^{past}$, while their future segment constitutes $\bm{Y}^{future}$.

	The goal is to predict the future target trajectory $\widetilde{\bm{Y}}^{target} \in \mathbb{R}^{C \times F}$ conditioned on $\bm{X}^{target}$, $\bm{X}^{past}$, and $\bm{Y}^{future}$.
	Formally, we learn a forecasting model $\mathcal{F}_\theta$ parameterized by $\theta$:
	\begin{equation}
		\widetilde{\bm{Y}}^{target} = \mathcal{F}_\theta\big(\bm{X}^{target}, \bm{X}^{past}, \bm{Y}^{future}\big).
	\end{equation}
	
	\subsection{Structure Overview}
	\begin{figure}[htbp]
		\begin{center}
			\centerline{\includegraphics[width=\columnwidth]{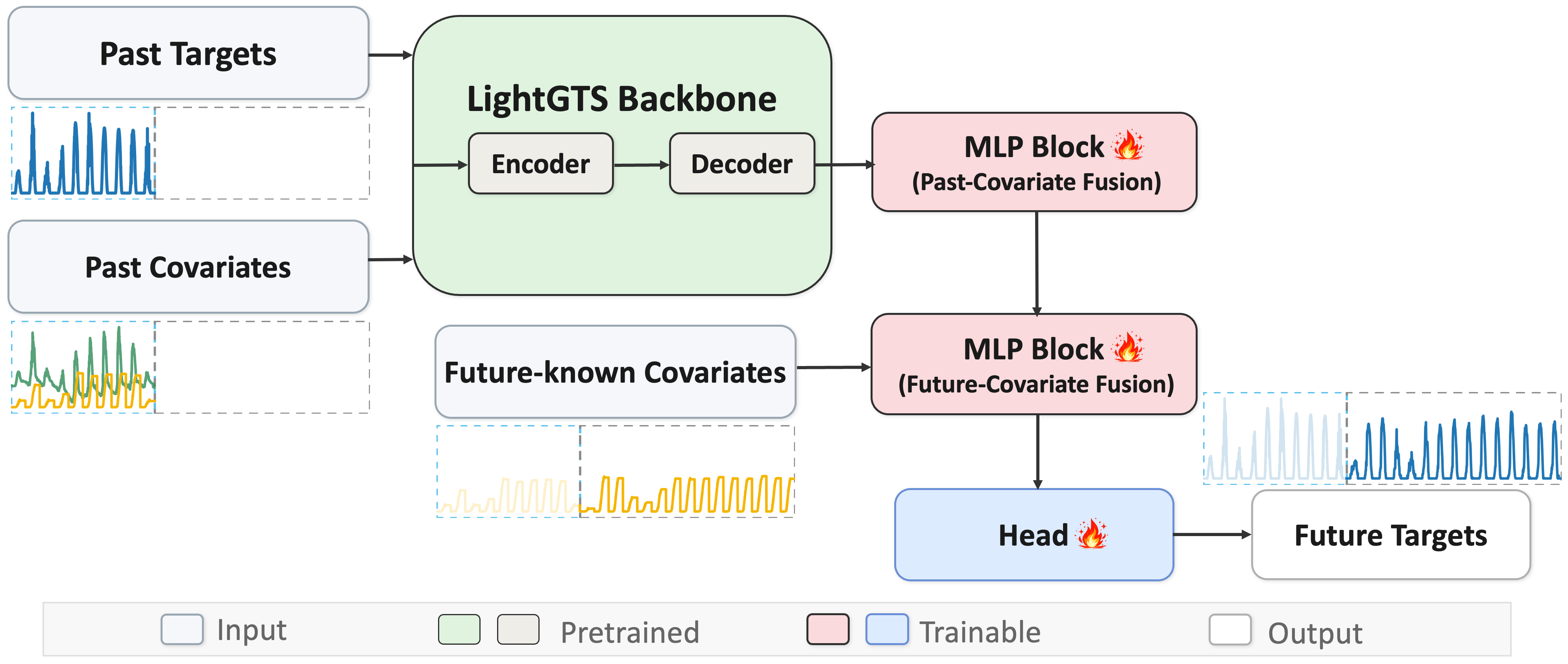}}
			\caption{Architecture of LightGTS-Cov. It takes as input past targets, past covariates, and future-known covariates, and consists of (i) a pretrained LightGTS backbone, (ii) a two-stage post-decoder MLP fusion module for past and future covariates, and (iii) a shared output head that projects tokens to the final forecast.} 
			\label{architecture}
		\end{center}
	\end{figure}
	
	Figure~\ref{architecture} illustrates the overall architecture of LightGTS-Cov.
	Our method builds upon the LightGTS~\cite{lightgts} backbone and adds lightweight post-decoder covariate fusion modules while preserving the original period-aware forecasting pipeline.
	Concretely, we reuse the periodical patching, flex projection, Transformer encoder--decoder, and periodical parallel decoding of LightGTS to obtain decoder-side target tokens $\bm{Z}^{target}$.
	In parallel, past covariates are processed with the same period-aware tokenization to produce aligned past covariate representations $\bm{Z}^{past}$ at the same token resolution.
	These aligned representations enable post-decoder, time-aligned covariate fusion without modifying the backbone decoding procedure.

	LightGTS-Cov performs post-decoder covariate integration in two stages.
	First, past covariates are encoded using the same period-aware tokenization as the target, and $\mathrm{MLP}_{past}$ performs past covariate fusion to produce intermediate tokens.
	Second, future-known covariates are embedded and aligned to the forecasting horizon, and $\mathrm{MLP}_{future}$ performs future covariate fusion to produce covariate-refined tokens $\bm{R}^{tile}$.
	Finally, we apply the shared output head to both $\bm{Z}^{target}$ and $\bm{R}^{tile}$ and sum the two projections to yield $\widetilde{\bm{Y}}^{target}$.

	\subsection{Target Backbone Representations}
	Given $\bm{X}^{target} \in \mathbb{R}^{C \times T}$, LightGTS applies periodical patching with a cycle length of $P$ and a flexible projection input layer to obtain encoder tokens.
	After the Transformer encoder and periodical parallel decoding, we obtain latent decoder representations for each target variable $i\in\{1,\dots,C\}$:
	\begin{equation}
		\bm{Z}^{target}_i=\{\bm{z}^{target}_{i,1},\dots,\bm{z}^{target}_{i,O}\}\in\mathbb{R}^{D\times O},
	\end{equation}
	where $D$ is the model dimension and $O=\lceil F/P\rceil$ is the number of decoded tokens covering the forecasting horizon. Stacking over all target variables yields $\bm{Z}^{target}\in\mathbb{R}^{C\times D\times O}$.

	\subsection{Covariate Pathways}
	\noindent\textbf{Past covariates.}
	For simplicity and to reuse a shared backbone, we encode past covariates $\bm{X}^{past}\in\mathbb{R}^{M_p\times T}$ using the same periodical patching scheme (cycle length $P$) and the same input projection as the target series. Passing these tokens through the backbone yields token-level latent representations:
	\begin{equation}
		\bm{Z}^{past}_m=\{\bm{z}^{past}_{m,1},\dots,\bm{z}^{past}_{m,O}\}\in\mathbb{R}^{D\times O},\quad m=1,\dots,M_p,
	\end{equation}
	which are stacked as $\bm{Z}^{past}\in\mathbb{R}^{M_p\times D\times O}$.
	With this shared backbone, $\bm{Z}^{past}$ is naturally time-aligned with $\bm{Z}^{target}$ at the token resolution, i.e., along the token index $j=1,\dots,O$.
	
	\noindent\textbf{Future-known covariates.}
	Future-known covariates $\bm{Y}^{future}\in\mathbb{R}^{M_f\times F}$ are available over the forecasting horizon.
	We tokenize them via patching, where the patch length $P^\ast$ can either follow the intrinsic periodicity of the covariate series (when known)  or be set to a fixed value for simplicity and consistency across datasets.
	This yields patched covariate tensors $\bm{Y}^{P^\ast}\in\mathbb{R}^{M_f\times P^\ast \times O^\ast}$, where $O^\ast=\lceil F/P^\ast\rceil$ denotes the number of patches covering the horizon.
	Each covariate patch is embedded into the backbone latent space via a linear embedding $W_f\in\mathbb{R}^{D\times P^\ast}$:
	\begin{equation}
		\bm{f}_{m,j}=W_f\,\bm{Y}^{P^\ast}_{m,:,j}\in\mathbb{R}^{D},\quad m=1,\dots,M_f,\ \ j=1,\dots,O.
	\end{equation}
	Stacking over all covariates and tokens yields $\bm{f}\in\mathbb{R}^{M_f\times D\times O}$, which is aligned to the forecasting horizon at the same token resolution.
	
	\subsection{Post-Decoder Fusion}
	\textbf{Two-stage MLP fusion.}
	LightGTS-Cov refines the decoder-side target tokens through two lightweight MLP blocks, $\mathrm{MLP}_{past}$ and $\mathrm{MLP}_{future}$.
	The two blocks share the same architecture (depth, hidden size, and activation) but use separate parameters to accommodate different input dimensions.
	For each forecast token index $j\in\{1,\dots,O\}$, we compute:
	
	\textbf{Stage 1: past-covariate fusion.}
	We concatenate the decoded target and past-covariate representations along the variable dimension:
	\begin{equation}
		\bm{Z}^{mix}_{:,:,j}=\big[\bm{Z}^{target}_{:,:,j}\,\Vert\,\bm{Z}^{past}_{:,:,j}\big]\in\mathbb{R}^{(C+M_p)\times D}.
	\end{equation}
	This design is motivated by the fact that each token index $j$ corresponds to the same period-aligned temporal segment across all variables.
	Concatenating along the variable dimension therefore preserves token-wise temporal alignment while exposing cross-variable evidence to the fusion module in a single view. This is in line with recent findings that modeling cross-variate interactions explicitly—e.g., by treating variates as the token axis—can be beneficial for multivariate forecasting~\citep{itransformer,wang2024timexer}. 
	
	We then flatten the concatenated matrix into a vector
	\begin{equation}
		\bm{u}^{(1)}_j=\mathrm{Flatten}\!\big(\bm{Z}^{mix}_{:,:,j}\big)\in\mathbb{R}^{(C+M_p)D},
	\end{equation}
	so that $\mathrm{MLP}_{past}$ can implement a flexible, token-wise mixing function over variable-specific latent features without introducing additional attention blocks or altering the backbone structure.
	The first MLP produces an intermediate token representation:
	\begin{equation}
		\bm{h}_j=\mathrm{MLP}_{past}\!\big(\mathrm{Norm}(\bm{u}^{(1)}_j)\big)\in\mathbb{R}^{D}.
	\end{equation}
	here, $\mathrm{Norm}(\cdot)$ denotes LayerNorm~\cite{Layernormalization}. Notably, $\bm{h}_j$ acts as a compact token-level summary of how past covariates refine the decoder token at index $j$, which also keeps the subsequent fusion with future-known covariates computationally light.
	
	\textbf{Stage 2: future-covariate fusion.}
	Future-known covariates provide horizon-aligned information for the forecast segment covered by token $j$ (e.g., weather forecasts or dispatch plans), and are widely recognized as critical signals in operational forecasting when they are available at prediction time~\cite{lim2021temporal, qiu2025dag}.
	We therefore inject future-known covariates by concatenating the intermediate token $\bm{h}_j$ with the future covariate embeddings at the same token index:
	\begin{equation}
		\bm{u}^{(2)}_j=\big[\bm{h}_j\,\Vert\,\bm{f}_{1,j}\,\Vert\,\cdots\,\Vert\,\bm{f}_{M_f,j}\big]\in\mathbb{R}^{(1+M_f)D},
	\end{equation}
	This design injects exogenous information at the output decoding stage, where tokens are already aligned with the forecast horizon, enabling the module to translate known future inputs into a direct residual correction of the predicted trajectory. 
	
	The second MLP then outputs a covariate-refined token:
	\begin{equation}
		\bm{r}_j=\mathrm{MLP}_{future}\!\big(\mathrm{Norm}(\bm{u}^{(2)}_j)\big)\in\mathbb{R}^{D}.
	\end{equation}
	Stacking $\{\bm{r}_j\}_{j=1}^O$ yields $\bm{R}\in\mathbb{R}^{D\times O}$.
	Since these future-known covariates are typically shared, system-level drivers (e.g., station-level weather forecasts or market-level dispatch signals) that affect all target variables coherently, we apply a shared refinement across targets and tile $\bm{R}$ along the variable dimension to match the backbone token shape:
	\begin{equation}
		\bm{R}^{tile}\in\mathbb{R}^{C\times D\times O},\qquad 
		\bm{R}^{tile}_{i,:,:}=\bm{R},\ \ i=1,\dots,C.
	\end{equation}
	\noindent\textbf{Forecast.}
	Finally, we project both the decoded target tokens $\bm{Z}^{target}$ and the tiled covariate-refined tokens $\bm{R}^{tile}$ to the data space using the \emph{same} LightGTS output head and obtain the final forecast through an additive composition:
	\begin{equation}
		\widetilde{\bm{Y}}^{target}
		= \mathrm{Head}\!\left(\bm{Z}^{target}\right) + \mathrm{Head}\!\left(\bm{R}^{tile}\right).
	\end{equation}
	This formulation performs residual refinement on top of the backbone:
	$\mathrm{Head}(\bm{Z}^{target})$ provides the original period-aware forecast, while $\mathrm{Head}(\bm{R}^{tile})$ supplies a covariate-conditioned correction.
	By placing covariate injection in an additive residual branch, the model can retain the backbone's forecasting behavior when covariates are absent or uninformative and only introduce controlled adjustments when covariates provide useful guidance.

	\subsection{Training Strategy}
	By default, we keep the pretrained LightGTS backbone fixed and train the added covariate modules together with the shared output head.
	Concretely, we optimize the future-covariate embedding matrix $W_f$, the parameters of $\mathrm{MLP}_{past}$ and $\mathrm{MLP}_{future}$, as well as the output head $\mathrm{Head}(\cdot)$ used for projections.
	This strategy keeps the trainable parameter set lightweight while enabling effective covariate integration.
	We also consider a full fine-tuning variant where the backbone is unfrozen and updated jointly with the covariate modules and $\mathrm{Head}(\cdot)$ for stronger task specialization.

	We adopt the mean squared error (MSE) over all target variables and forecast steps.
	Given ground-truth future targets $\bm{Y}^{target}$ and predictions $\widetilde{\bm{Y}}^{target}$, the loss is
	\begin{equation}
		\mathcal{L}_{\mathrm{MSE}}=\big\|\bm{Y}^{target}-\widetilde{\bm{Y}}^{target}\big\|_F^2,
	\end{equation}
	where $\|\cdot\|_F$ denotes the Frobenius norm.

	\section{Experiments}
	\label{sec:experiments}

	\subsection{Experimental Setup}
	
	\noindent\textbf{Datasets.}
	To evaluate covariate-enhanced forecasting, we build an evaluation suite spanning public benchmarks and two industrial deployments. Each dataset provides a target series together with historical covariates and (when available) horizon-aligned future-known covariates. Our public evaluation covers six real-world datasets: five electricity price forecasting tasks from the EPF benchmark~\cite{lago2021forecasting} and one multi-source generation dataset (Energy)~\cite{qiu2025dag}. We further validate practical effectiveness in two operational settings with an industry partner: (i) long-term PV power forecasting for a multi-inverter station (15 days, 5-minute resolution) using on-site sensor histories and future weather forecasts, and (ii) day-ahead electricity price forecasting (1 day, 15-minute resolution) using historical prices with future-known dispatch disclosures and meteorological forecasts. Detailed dataset statistics and descriptions are provided in Appendix~\ref{sec:A1} (Table~\ref{dataset_stats}). Further implementation details can be found in the Appendix~\ref{sec:A2} and~\ref{sec:A3}.
	
	\noindent\textbf{Baselines.}
	We compare LightGTS-Cov against a broad set of baselines in two categories:
	(1) Covariate-aware deep forecasting models trained from scratch. This group includes covariate-aware models with native support for future-known covariates, such as TFT~\cite{lim2021temporal}, TimeXer~\cite{wang2024timexer}, and TiDE~\cite{das2023long}, as well as models that do not natively handle future covariates, including PatchTST~\cite{PatchTST}, iTransformer~\cite{itransformer}, CrossLinear~\cite{zhou2025crosslinear}, and DLinear~\cite{DLinear}. Following the setup in~\cite{qiu2025dag}, we augment the latter with an MLP-based fusion module to incorporate future-known covariates.
	(2) Adaptation methods for TSFMs. These methods attach lightweight plug-ins to pretrained backbones to incorporate covariates with limited additional parameters, including ChronosX IIB/OIB~\cite{arango2025chronosx} for Chronos, CoRA~\cite{qin2025cora} for Sundial, UniCA~\cite{han2025unica} for TimesFM, and AdaPTS~\cite{benechehab2025adapts} for MOMENT.
	Timer-XL~\cite{timerxl} incorporates covariate-informed contexts via its TimeAttention mechanism.
	
	\subsection{Results on Public Benchmarks}\label{plugin}
	We evaluate LightGTS-Cov on covariate-rich electricity and energy datasets, including EPF and Energy.
	In EPF, the day-ahead electricity price is the target series, and two associated series are used as covariates.
	For Energy, we forecast one target series given five associated covariate series.
	Following the standard protocol, we fix the lookback and forecast horizons to 168 and 24 time steps, respectively.
	
	\noindent\textbf{Comparison with deep forecasting models.}
	Firstly, we compare LightGTS-Cov with representative end-to-end deep forecasters trained under full-parameter supervision and report two settings: historical-only inputs (Table~\ref{table2}) and additionally providing future-known covariates over the forecasting horizon (Table~\ref{table3}). Overall, removing future-known covariates causes a consistent performance drop across datasets, underscoring that horizon-aligned exogenous signals are important for accurate and stable day-ahead forecasts; 
	nevertheless, LightGTS-Cov remains highly competitive in the historical-only setting, achieving the best or tied-best performance on most datasets and staying very close to the strongest results on the remaining ones. 
	When future-known covariates are available, most baselines improve, and LightGTS-Cov matches the performance of leading methods in the field and attains the best results on several benchmarks under both MSE and MAE.
	This highlights the benefit of token-aligned covariate representations and post-decoder residual refinement that injects exogenous information without disrupting the backbone decoding pipeline.

	\begin{table*}[h]  
		\caption{Full results on public benchmarks without future-known covariates. Results of baseline models are taken from DAG~\cite{qiu2025dag}. \textcolor{red}{Red}: best. \underline{\textcolor{blue}{Blue}}: second best.}
		\label{table2}
		\centering
		\resizebox{0.95\textwidth}{!}{%		
			\begin{tabular}{lcccccccccccccccc}
				\toprule
				Models &
				\multicolumn{2}{c}{LightGTS-Cov} &
				\multicolumn{2}{c}{DAG} &
				\multicolumn{2}{c}{TimeXer} &
				\multicolumn{2}{c}{TFT} &
				\multicolumn{2}{c}{TiDE} &
				\multicolumn{2}{c}{CrossLinear} &
				\multicolumn{2}{c}{PatchTST} &
				\multicolumn{2}{c}{DLinear} 
				\\
				\cmidrule(lr){2-3}\cmidrule(lr){4-5}\cmidrule(lr){6-7}
				\cmidrule(lr){8-9}\cmidrule(lr){10-11}\cmidrule(lr){12-13}\cmidrule(lr){14-15}\cmidrule(lr){16-17}
				Metric & MSE & MAE & MSE & MAE & MSE & MAE & MSE & MAE & MSE & MAE & MSE & MAE & MSE & MAE & MSE & MAE \\
				\midrule
				NP  & \best{0.221} & \best{0.259} & \secondbest{0.237} & \best{0.259} & 0.270 & 0.292 & 0.330 & 0.321 & 0.297 & 0.309 & 0.246 & \secondbest{0.282} & 0.283 & 0.296 & 0.294 & 0.308  \\
				PJM & \best{0.072} & \secondbest{0.168}  & \best{0.072} & \best{0.162} & \secondbest{0.096} & 0.191 & 0.129 & 0.231 & 0.105 & 0.213 & \secondbest{0.096} & 0.198 & 0.106 & 0.212 & 0.109 & 0.215 \\
				BE  & \best{0.365} & \best{0.243} & \secondbest{0.376} & \secondbest{0.244} & 0.392 & 0.253 & 0.400 & 0.267 & 0.492 & 0.322 & 0.389 & 0.253 & 0.405 & 0.264 & 0.453 & 0.308 \\
				FR  & 0.486 & 0.198 & \secondbest{0.366} & \secondbest{0.195} & 0.445 & 0.231 & \best{0.345} & \best{0.190} & 0.415 & 0.254 & 0.397 & 0.204 & 0.397 & 0.223 & 0.412 & 0.253  \\
				DE  & \best{0.414} & \best{0.399} & \secondbest{0.422} & \secondbest{0.411} & 0.501 & 0.445 & 0.576 & 0.452 & 0.465 & 0.433 & 0.438 & 0.423 & 0.503 & 0.450 & 0.511 & 0.458  \\
				Energy  & \best{0.100} & \best{0.243}& 0.112 & 0.257 & 0.138 & 0.293 & 0.352 & 0.463 & 0.117 & 0.265 & \secondbest{0.102} & \secondbest{0.246} & 0.108 & 0.254 & 0.106 & 0.248  \\
				\bottomrule
			\end{tabular}
		}
	\end{table*}

	\begin{table*}[htbp]   
		\caption{Full results on public benchmarks with future-known covariates. Results of baseline models are taken from DAG~\cite{qiu2025dag}. \textcolor{red}{Red}: best. \underline{\textcolor{blue}{Blue}}: second best.}
		\label{table3}
		\centering
		\resizebox{0.95\textwidth}{!}{%		
			\begin{tabular}{lcccccccccccccccc}
				\toprule
				Models &
				\multicolumn{2}{c}{LightGTS-Cov} &
				\multicolumn{2}{c}{DAG} &
				\multicolumn{2}{c}{TimeXer} &
				\multicolumn{2}{c}{TFT} &
				\multicolumn{2}{c}{TiDE} &
				\multicolumn{2}{c}{CrossLinear} &
				\multicolumn{2}{c}{PatchTST} &
				\multicolumn{2}{c}{DLinear} \\
				\cmidrule(lr){2-3}\cmidrule(lr){4-5}\cmidrule(lr){6-7}\cmidrule(lr){8-9}\cmidrule(lr){10-11}\cmidrule(lr){12-13}\cmidrule(lr){14-15}\cmidrule(lr){16-17}
				Metric & MSE & MAE & MSE & MAE & MSE & MAE & MSE & MAE & MSE & MAE & MSE & MAE & MSE & MAE & MSE & MAE \\
				\midrule
				NP  & \best{0.167} & \best{0.216} & \secondbest{0.202} & \secondbest{0.237} & 0.236 & 0.266 & 0.219 & 0.249 & 0.284 & 0.301 & 0.210 & 0.266 & 0.249 & 0.294 & 0.282 & 0.337  \\
				PJM & \secondbest{0.070} & \secondbest{0.157} & \best{0.057} & \best{0.143} & 0.075 & 0.166 & 0.095 & 0.195 & 0.106 & 0.214 & 0.088 & 0.191 & 0.116 & 0.239 & 0.096 & 0.208  \\
				BE  & \best{0.338} & \best{0.224} & \secondbest{0.361} & \secondbest{0.229} & 0.392 & 0.253 & 0.426 & 0.272 & 0.426 & 0.285 & 0.391 & 0.259 & 0.452 & 0.326 & 0.451 & 0.310  \\
				FR  & \best{0.347} & \secondbest{0.177} &\secondbest{0.355} & \best{0.171} & 0.366 & 0.208 & 0.543 & 0.253 & 0.418 & 0.255 & 0.390 & 0.226 & 0.518 & 0.368 & 0.422 & 0.265  \\
				DE  & \secondbest{0.289} & \secondbest{0.331} & \best{0.277} & \best{0.322} & 0.339 & 0.362 & 0.380 & 0.383 & 0.367 & 0.383 & 0.479 & 0.387 & 0.396 & 0.412 & 0.423 & 0.426  \\
				Energy  & \best{0.079} & \secondbest{0.218} &\best{0.079} & \best{0.215} & 0.122 & 0.273 & 0.093 & 0.235 & 0.103 & 0.248 & 0.241 & 0.418 & 0.239 & 0.390 & \secondbest{0.085} & 0.230 \\
				\bottomrule
			\end{tabular}
		}
	\end{table*}

	\noindent\textbf{Comparison with adaptation methods.}
	For a comprehensive comparison, we benchmark LightGTS-Cov against representative plug-in based TSFM adaptation methods, including Sundial+CoRA, MOMENT+AdaPTS, TimesFM+UniCA, and Chronos with IIB/OIB-style adapters.
	For each adapter, we follow the original papers and pair it with its reported backbone and configuration.
	We further include Timer-XL with TimeAttention as an additional covariate-aware plug-in baseline.
	Table~\ref{table1} reports results under the future-known covariate setting, where LightGTS-Cov achieves consistently low errors across EPF markets compared with these adapter-based baselines.
	Moreover, Table~\ref{table4} presents controlled ablations on the same LightGTS backbone by attaching different plug-ins (CoRA, IIB, OIB, and their combinations).
	Relative to vanilla LightGTS, the proposed Cov plug-in yields a substantial reduction in both MSE and MAE (around 28\% improvement on average), demonstrating that effective covariate conditioning brings clear benefits beyond the base model.
	It also remains competitive against alternative adapter designs, suggesting strong compatibility with the LightGTS architecture in covariate-aware forecasting.
	
	\begin{table*}[ht]  
		\caption{Full results on public benchmarks with future-known covariates using pretrained TSFMs and adaptation methods. Results of adapter-based baselines are taken from CoRA~\cite{qin2025cora}. \textcolor{red}{Red}: best. \underline{\textcolor{blue}{Blue}}: second best.}
		\label{table1}
		\centering
		\resizebox{0.8\textwidth}{!}{%	
			\begin{tabular}{lcccccccccccccc}
				\toprule
				Models &
				\multicolumn{2}{c}{LightGTS-Cov} &
				\multicolumn{2}{c}{AdaPTS} &
				\multicolumn{2}{c}{UniCA} &
				\multicolumn{2}{c}{ChronosX} &
				\multicolumn{2}{c}{CoRA} &
				\multicolumn{2}{c}{Timer-XL} \\
				\cmidrule(lr){2-3}\cmidrule(lr){4-5}\cmidrule(lr){6-7}	\cmidrule(lr){8-9}\cmidrule(lr){10-11}\cmidrule(lr){12-13}
				Metric & MSE & MAE & MSE & MAE & MSE & MAE & MSE & MAE & MSE & MAE & MSE & MAE  \\
				\midrule
				NP  & \best{0.167} & \best{0.216} & 0.231 & 0.259 & 0.265 & 0.289 & 0.254 & 0.278 & \secondbest{0.222} & \secondbest{0.246} & 0.234 & 0.262 \\
				PJM & \best{0.070} & \best{0.157} & 0.080 & 0.173 & 0.090 & 0.187 & 0.089 & 0.189 & \secondbest{0.073} & \secondbest{0.165} & 0.089 & 0.187 \\
				BE  & \best{0.338} & \best{0.224} & 0.355 & 0.261 & 0.368 & 0.273 & 0.371 & 0.274 & \secondbest{0.339} & \secondbest{0.236} & 0.371 & 0.243 \\
				FR  & \best{0.347} & \best{0.177} & 0.363 & 0.218 & 0.365 & 0.218 & 0.361 & 0.217 & \secondbest{0.357} & 0.206 & 0.381 & \secondbest{0.204} \\
				DE  & \best{0.289} & \best{0.331} & 0.455 & 0.424 & 0.553 & 0.466 & 0.453 & 0.426 & \secondbest{0.401} & \secondbest{0.388} & 0.434 & 0.415 \\

				\bottomrule
		\end{tabular}}
	\end{table*}

	\begin{table*}[ht]  
		\caption{Performance of LightGTS with diverse covariate adapters in covariate-aware forecasting with future-known covariates. \textcolor{red}{Red}: best. \underline{\textcolor{blue}{Blue}}: second best.}
		\label{table4}
		\centering
		\resizebox{0.9\textwidth}{!}{%		
			\begin{tabular}{lcccccccccccccc}
				\toprule
				Models &
				\multicolumn{2}{c}{LightGTS} &
				\multicolumn{2}{c}{LightGTS+Cov} &
				\multicolumn{2}{c}{LightGTS+CoRA} &
				\multicolumn{2}{c}{LightGTS+IIB} &
				\multicolumn{2}{c}{LightGTS+OIB} &
				\multicolumn{2}{c}{LightGTS+IIB+OIB} \\
				\cmidrule(lr){2-3}\cmidrule(lr){4-5}\cmidrule(lr){6-7}	\cmidrule(lr){8-9}\cmidrule(lr){10-11}\cmidrule(lr){12-13}
				Metric & MSE & MAE & MSE & MAE & MSE & MAE & MSE & MAE & MSE & MAE & MSE & MAE  \\
				\midrule
				NP  & 0.248 & 0.271 & \best{0.167} & \best{0.216} & 0.184 & 0.232 & 0.224 & 0.263 & 0.182 & \secondbest{0.223} & \secondbest{0.176} & \secondbest{0.223} \\
				PJM & 0.098 & 0.198 & \secondbest{0.070} & \secondbest{0.157} & \best{0.062} & \secondbest{0.157} & 0.071 & 0.165 & 0.078 & \secondbest{0.157} & \best{0.062} & \best{0.150} \\
				BE  & 0.392 & 0.254 & \best{0.338} & \best{0.224} & 0.358 & 0.241 & \secondbest{0.355} & 0.241 & 0.379 & 0.242 & 0.357 & \secondbest{0.231} \\
				FR  & 0.414 & 0.215 & \best{0.347} & \best{0.177} & 0.397 & 0.193 & \secondbest{0.353} & \secondbest{0.189} & 0.416 & 0.196 & 0.366 & 0.190 \\
				DE  & 0.446 & 0.417 & \best{0.289} & \best{0.331} & 0.316 & 0.342 & 0.455 & 0.409 & \secondbest{0.297} & 0.340 & 0.306 & \secondbest{0.336} \\
				Energy & 0.108 & 0.252 & \best{0.079} & \best{0.218} & \secondbest{0.090} & \secondbest{0.230} & 0.103 & 0.247 & 0.103 & 0.251 & 0.096 & 0.241 \\

				\bottomrule
		\end{tabular}}
		
	\end{table*}

	\begin{table*}[!t]  
		\caption{Parameter sensitivity analysis of the covariate MLP block under future-known covariates with respect to the hidden dimension $D_h$ and depth $L$. \textcolor{red}{Red}: best. \underline{\textcolor{blue}{Blue}}: second best.}
		\label{table5}
		\centering
		\resizebox{\textwidth}{!}{%	
			\begin{tabular}{lcccccccc|cccccccc}
				\toprule
				Models &
				\multicolumn{2}{c}{$D_h = D/4$} &
				\multicolumn{2}{c}{$D_h = D/2$} &
				\multicolumn{2}{c}{$D_h = 2D$} &
				\multicolumn{2}{c}{$D_h = 4D$} &
				\multicolumn{2}{c}{$L = 1$} &
				\multicolumn{2}{c}{$L = 2$} &
				\multicolumn{2}{c}{$L = 3$} &
				\multicolumn{2}{c}{$L = 4$} \\
				\cmidrule(lr){2-3}\cmidrule(lr){4-5}\cmidrule(lr){6-7}\cmidrule(lr){8-9}\cmidrule(lr){10-11}\cmidrule(lr){12-13}\cmidrule(lr){14-15}\cmidrule(lr){16-17}
				Metric & MSE & MAE & MSE & MAE & MSE & MAE & MSE & MAE  & MSE & MAE & MSE & MAE & MSE & MAE & MSE & MAE \\
				\midrule
				NP      & \best{0.167} & \best{0.216} & \secondbest{0.169} & \secondbest{0.217} & 0.171 & 0.219 & 0.174 & 0.220 & 0.173 & \best{0.216} & 0.173 & 0.219 & 0.173 & 0.219 & 0.175 & 0.221 \\
				
				PJM     & \secondbest{0.070} & 0.157 & 0.073 & \secondbest{0.151} & 0.071 & 0.152 & \best{0.065} & \best{0.148} & 0.071 & 0.153 & \secondbest{0.070} & \secondbest{0.151} & 0.082 & 0.153 & 0.093 & 0.152 \\
				
				BE      & \best{0.338} & \best{0.224} & 0.344 & 0.230 & 0.348 & 0.236 & 0.352 & 0.236  & \secondbest{0.343} & 0.229 & 0.348 & 0.231 & 0.348 & \secondbest{0.227} & 0.352 & 0.231 \\
				
				FR      & \secondbest{0.347} & \best{0.177} & \secondbest{0.347} & 0.181 & \best{0.343} & \secondbest{0.180} & 0.360 & 0.186  & 0.373 & 0.196 & 0.348 & 0.182 & 0.375 & 0.184 & 0.377 & 0.188 \\
				
				DE      & 0.289 & 0.331 & 0.286 & \secondbest{0.329} & \best{0.281} & 0.331 & \secondbest{0.282} & \secondbest{0.329} & 0.289 & 0.331 & 0.289 & \best{0.328} & 0.284 & 0.336 & 0.287 & 0.332 \\
				
				Energy  & 0.086 & 0.227 & \secondbest{0.085} & \secondbest{0.226} & 0.091 & 0.236 & \secondbest{0.085} & 0.228 & 0.086 & 0.227 & \best{0.082} & \best{0.224} & 0.086 & 0.227 & 0.087 & 0.229 \\
				\bottomrule
				
		\end{tabular}}
	\end{table*}
	
	\noindent\textbf{Parameter sensitivity.}
	We analyze the sensitivity of the plug-in MLP by varying the hidden dimension $D_h$ and the number of layers $L$.
	As shown in Table~\ref{table5}, performance remains stable across a wide range of configurations, with only minor fluctuations in MSE/MAE.
	This suggests that LightGTS-Cov is not overly sensitive to the specific MLP capacity, and the improvements are primarily driven by effective covariate conditioning rather than simply increasing plug-in parameters.

	% ---- PV active power forecasting ----
	\begin{figure}[ht]
		\centering
		\includegraphics[width=\linewidth]{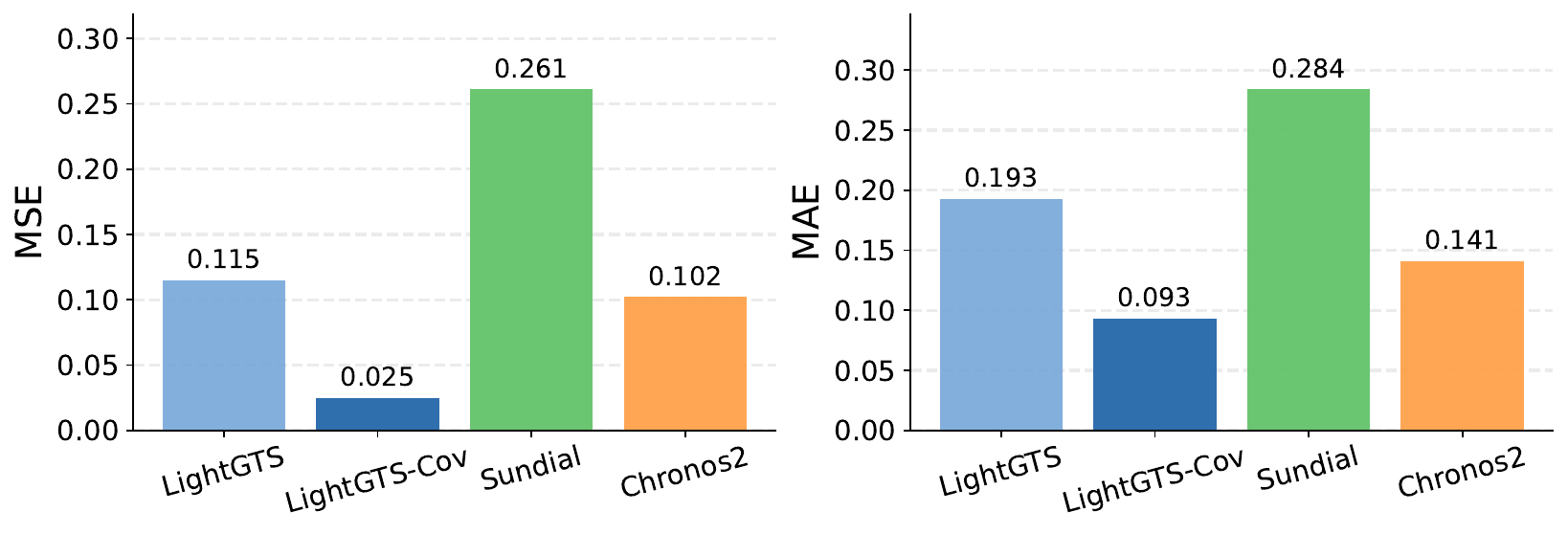}
		\caption{Long-term PV power forecasting results in Industrial Deployment I: MSE (left) and MAE (right).}
		\label{fig:pv_active_power_bar}
	\end{figure}

	\begin{figure}[ht]
		\centering
		\includegraphics[width=\linewidth]{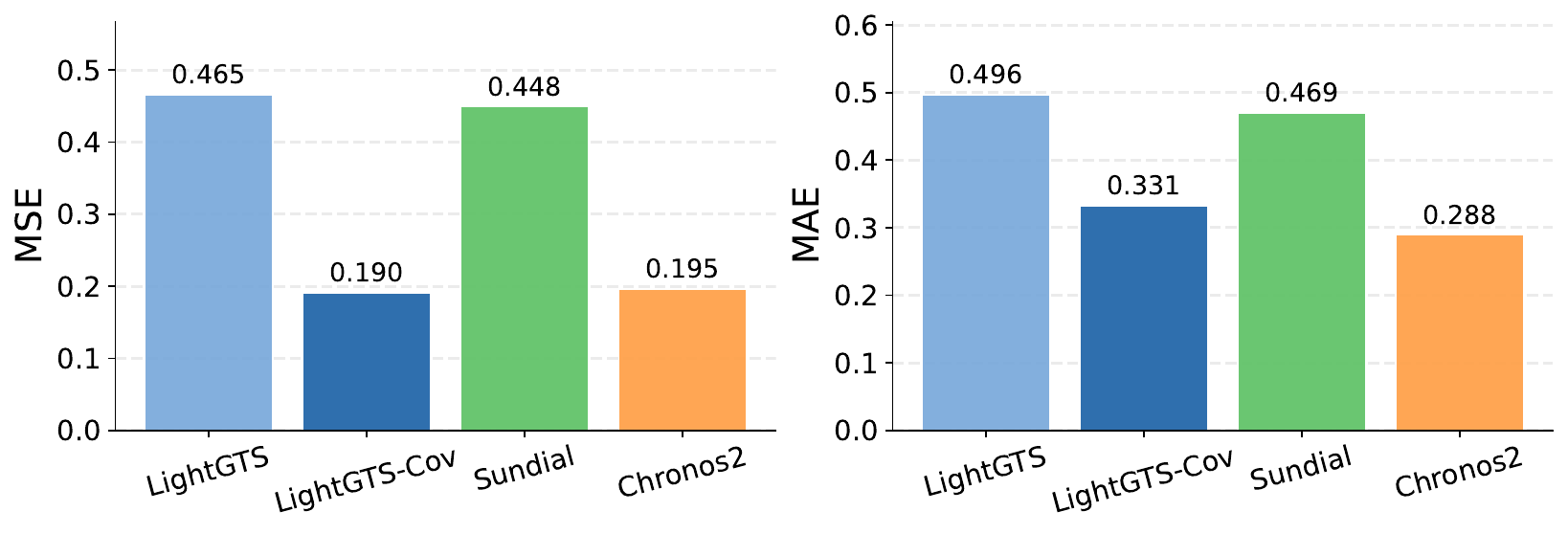}
		\caption{Day-ahead electricity price forecasting results in Industrial Deployment II: MSE (left) and MAE (right).}
		\label{fig:elec_price_bar}
	\end{figure}

	\subsection{Industrial Deployment I: Long-term PV Power Forecasting}
	
	\noindent\textbf{Background and motivation.}
	Accurate photovoltaic (PV) power forecasting is fundamental for day-ahead and multi-day operational planning, including dispatch scheduling, imbalance mitigation, and maintaining grid reliability~\cite{akhter2019review,gupta2021pv}. In practice, PV output is mainly driven by weather-especially solar irradiance---and follows strong diurnal/seasonal patterns. This makes exogenous inputs crucial for long-horizon forecasting, where uncertainty grows with lead time and autoregressive forecasts can drift or misestimate peak levels under changing conditions.
	
	\noindent\textbf{Data sources.}
	We study a real PV site and forecast device-level active power. The dataset combines three sources:
	\textbf{(i) inverter power} (23 channels) as targets;
	\textbf{(ii) on-site environmental sensors} (irradiance, temperature, humidity, wind) used only as \emph{past} covariates; and
	\textbf{(iii) weather forecasts} over the 15-day horizon (GHI, DNI, air temperature, relative humidity) as \emph{future-known} covariates.
	
	\noindent\textbf{Forecasting task.}
	The task is to predict the next 15 days of PV active power at 5-minute resolution for multiple devices within the same site. We adopt a direct multi-step paradigm: given the most recent 10 days of history, the model outputs the entire 15-day trajectory in a single forward pass. This formulation matches operational requirements and avoids error accumulation associated with iterative rollouts.
	
\begin{figure*}[ht]
		\begin{center}
			\centerline{\includegraphics[width=1.8\columnwidth]{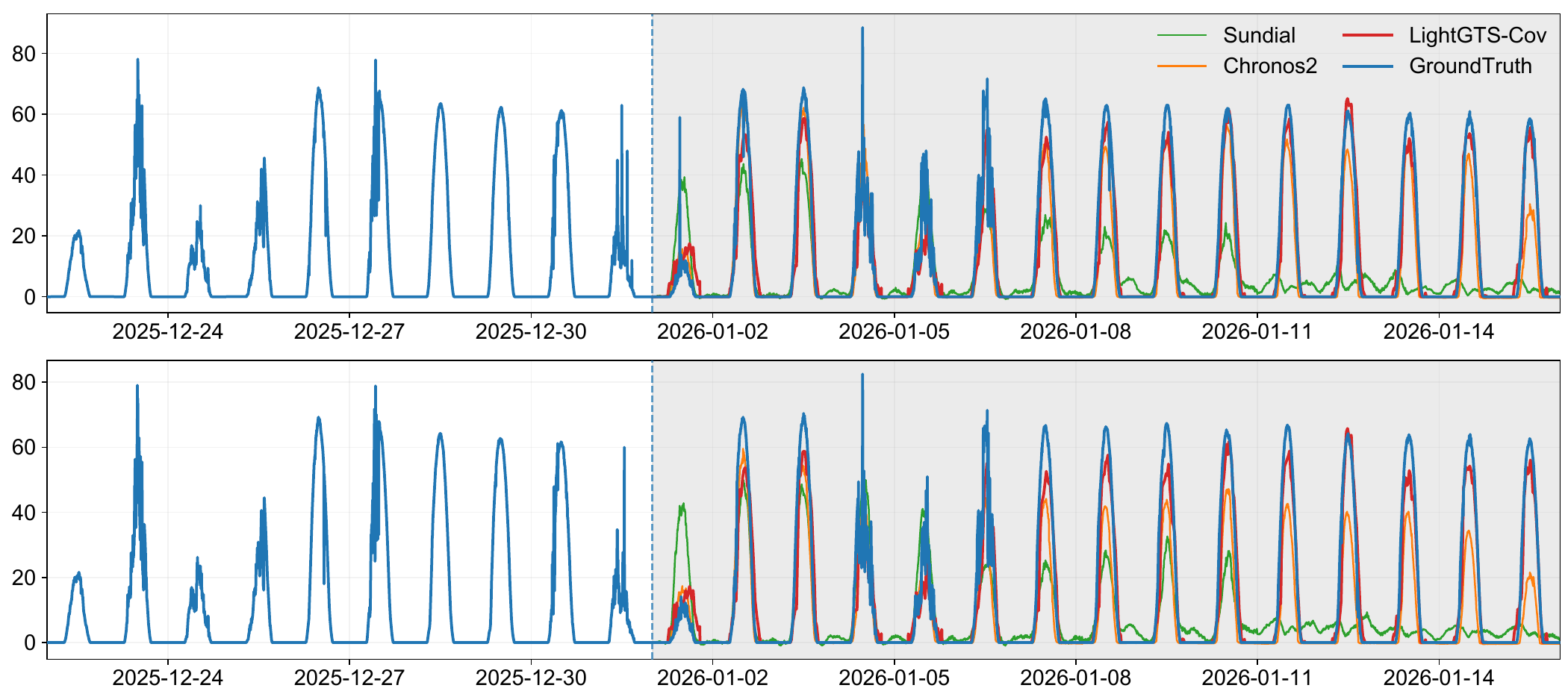}}
			\caption{PV power forecasting comparison over two inverter channels (top: \#17; bottom: \#23) using 10-day history to predict the next 15 days at 5-minute resolution. The dashed line marks the forecast start. LightGTS-Cov better aligns peak timing and magnitude. Curves show ground truth (blue) and predictions from LightGTS-Cov (red), Chronos-2 (orange), and Sundial (green).}
			\label{result1}
		\end{center}
	\end{figure*}

	\noindent\textbf{Evaluation and results.}
	We compare LightGTS-Cov (1.1M parameters) with two representative foundation-model baselines under a unified evaluation protocol: Chronos-2~\cite{chronos2} (120M parameters), a recent state-of-the-art TSFM that extends Chronos toward universal forecasting with support for covariate-informed settings, and Sundial~\cite{liu2025sundial} (128M parameters), which adopts a flow-matching–based training objective via TimeFlow Loss to enable probabilistic forecasting and is suitable for industrial forecasting applications.
	Figure~\ref{fig:pv_active_power_bar} summarizes the overall forecasting performance in terms of MSE and MAE, including the LightGTS backbone as a reference. LightGTS-Cov yields a substantial improvement over vanilla LightGTS, confirming that the proposed covariate plug-in brings clear gains on top of the original backbone. Under the same covariate setting, LightGTS-Cov achieves the lowest errors on both metrics among all compared models, outperforming the much larger Chronos-2 and Sundial and demonstrating strong competitiveness in long-term PV power forecasting.
	To further examine qualitative behaviors, Figure~\ref{result1} visualizes predictions over representative windows for two inverter channels, plotting ground truth against forecasts from each model. While all methods recover the dominant diurnal structure, they differ notably in peak magnitudes and ramp-up/ramp-down dynamics. LightGTS-Cov tracks peak levels and intra-day transitions more faithfully, whereas Sundial shows larger deviations toward the mid-to-late horizon, consistent with error accumulation in long-term rollouts. Beyond accuracy, LightGTS-Cov is also deployment-oriented: with only 1M parameters, it supports low-latency inference (on the order of seconds in practice), making it suitable for industrial forecasting pipelines.

	% ---- Electricity price forecasting ----
	\begin{figure*}[!h]
		\begin{center}
			\centerline{\includegraphics[width=1.8\columnwidth]{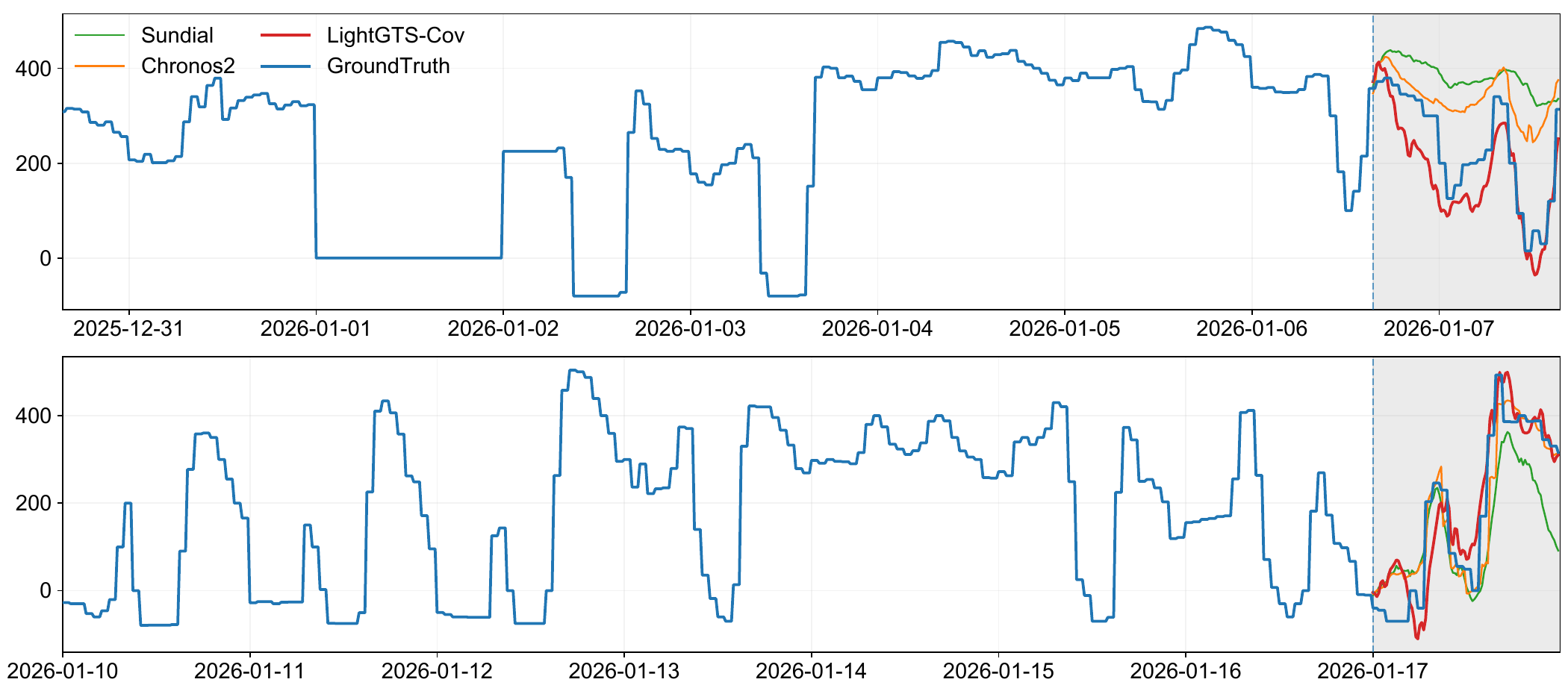}}
			\caption{Day-ahead electricity price forecasting under two representative windows (top: weekday; bottom: weekend). The dashed line marks the forecast start and the shaded region indicates the forecast horizon. LightGTS-Cov better tracks trend changes. Curves show ground truth (blue) and predictions from LightGTS-Cov (red), Chronos-2 (orange), and Sundial (green).}
			
			\label{result2}
		\end{center}
	\end{figure*}

	\subsection{Industrial Deployment II: Day-ahead Electricity Price Forecasting}

	\noindent\textbf{Background and motivation.}
	Day-ahead electricity price forecasting is a core capability in market-based power system operations, underpinning bidding and procurement decisions as well as portfolio risk management~\cite{weron2014electricity}. Day-ahead prices exhibit a pronounced intraday peak–valley structure and systematic calendar effects (weekday vs.\ weekend/holiday), while remaining highly sensitive to supply–demand tightness and uncertainty in renewable generation. This combination can induce abrupt regime shifts and, under surplus conditions, negative prices. These characteristics motivate forecasting models that integrate historical prices with exogenous drivers capturing demand, renewable supply, and meteorological conditions.
	
	\noindent\textbf{Data sources.}
	We study a regional day-ahead market and consider three categories of inputs:
	\textbf{(i) historical day-ahead prices} as the target series;
	\textbf{(ii) dispatch disclosures} summarizing next-day supply--demand plans (e.g., dispatched load, tie-line imports, and generation by type). We further construct a bidding-space indicator from these disclosed components, which is informative for price formation. As these plans are released around 10{:}00 (local time), we treat them as future-known covariates for next-day forecasting; and
	\textbf{(iii) weather forecasts} that affect both demand and renewable output and are treated as future-known covariates over the next-day horizon. 
	
	\noindent\textbf{Forecasting task.}
	The task is next-day day-ahead price forecasting at a 15-minute resolution: given a 7-day historical context window, the model predicts the full next-day price curve (24 hours) in a single forward pass. This formulation matches operational usage and avoids error accumulation from iterative rollouts.
	
	\noindent\textbf{Evaluation and results.}
	We compare LightGTS-Cov with representative foundation-model baselines (Chronos-2~\cite{chronos2} and Sundial~\cite{liu2025sundial}) under a unified evaluation protocol. Figure~\ref{fig:elec_price_bar} summarizes the overall performance in terms of MSE and MAE across four models. LightGTS-Cov achieves errors close to Chronos-2 and substantially improves upon the LightGTS backbone and Sundial, indicating that explicitly leveraging horizon-aligned future-known exogenous information is critical for operational day-ahead price forecasting. Figure~\ref{result2} further presents qualitative comparisons over representative weekday and weekend windows. LightGTS-Cov closely follows the ground truth and remains stable across different demand regimes; Chronos-2 also captures the main variations with competitive accuracy. In contrast, Sundial tends to generate overly smoothed trajectories and shows larger deviations around peaks and troughs, suggesting an underreaction to abrupt changes. Overall, jointly incorporating planned supply--demand signals from dispatch disclosures and weather-driven renewable factors improves both robustness and accuracy in day-ahead electricity price forecasting.
	
	\section{Conclusion}
	We propose \textbf{LightGTS-Cov}, a lightweight covariate-enhanced forecaster for production-oriented, covariate-rich settings. Built on the compact LightGTS backbone (\(\sim\)1M parameters), it maintains the original period-aware patching and parallel decoding, and adds a small decoder-side MLP plug-in (\(\sim\)0.1M parameters) that aligns past and future-known covariates to produce a residual refinement, thereby improving forecasts with minimal overhead.
	
	Across public benchmarks, LightGTS-Cov yields consistent gains with future-known covariates and remains competitive without them, demonstrating robustness to covariate availability. Compared with recent TSFM adaptation plug-ins and larger foundation-model baselines under matched covariate settings, LightGTS-Cov delivers consistently competitive accuracy, highlighting the effectiveness of the proposed decoder-side residual fusion. Finally, two industrial deployments validate its practical value: weather-forecast injection improves long-term PV power forecasting stability and peak tracking, while dispatch and meteorological covariates enhance day-ahead electricity price forecasting, especially around peak–valley dynamics.
	
	\noindent\textbf{Limitations and future work.}
	While LightGTS-Cov shows that a lightweight decoder-side residual plug-in can effectively leverage both past and future-known covariates, several directions remain. 
	First, the plug-in is largely backbone-agnostic and can be ported to other TSFMs by attaching a residual fusion branch at the representation-to-output interface. 
	Second, future-known covariates can be noisy or partially missing in practice; incorporating uncertainty-aware gating and explicit missingness modeling may further improve robustness under deployment shifts. 
	Finally, embedding stronger operational constraints (e.g., capacity bounds, or market rules) directly into the residual branch could enhance physical and economic consistency.

	\newpage
	{\footnotesize
	\bibliographystyle{ACM-Reference-Format}
	\bibliography{sample-base}}

	\newpage
	\appendix
	\section{Experimental Details}
	
	\begin{table*}[h]
		\caption{Dataset and key properties.}
		\label{dataset_stats}
		\centering
		\resizebox{\textwidth}{!}{%		
			\begin{tabular}{c c c c c c c c}
				\toprule
				Dataset & \#Targets & \#Covariates & Covariate Descriptions & Target Descriptions & Sampling Frequency & Length & Split \\
				\midrule
				NP   & 1 & 2 & Grid load, wind power  & Nord Pool electricity price & 1 Hour & 52,416 & 7:1:2 \\
				PJM  & 1& 2 & System load, COMED zonal load & PJM electricity price & 1 Hour & 52,416 & 7:1:2 \\
				BE   & 1& 2 & Generation, system load & Belgium electricity price & 1 Hour & 52,416 & 7:1:2 \\
				FR   & 1& 2 & Generation, system load & France electricity price & 1 Hour & 52,416 & 7:1:2 \\
				DE   & 1& 2 & Wind power, zonal load & Germany electricity price & 1 Hour & 52,416 & 7:1:2 \\
				Energy & 1 & 5 & Battery storage, wind, hydroelectric, solar, geothermal & Thermoelectric generation & 1 Hour & 13,064 & 7:1:2 \\
				PV power & 23 & 8 & On-site sensor measurements & Inverter-level active power & 5 Minutes & 61,633 & 7:1:2 \\
				Day-ahead electricity price & 1 & 22 & Dispatch disclosures, weather forecast & Day-ahead electricity price  & 15 Minutes & 37,057 & 7:1:2 \\
				\bottomrule
		\end{tabular}}
	\end{table*}
	
	\subsection{Datasets}\label{sec:A1}
	
	Table~\ref{dataset_stats} summarizes all datasets used in this work, including their sampling frequency, total length, and the numbers of target variables and covariates.
	
	\paragraph{\textbf{EPF} (five electricity markets).}
	We use the Electricity Price Forecasting (EPF) benchmark comprising five major markets: NP (Nord Pool), PJM, BE (Belgium), FR (France), and DE (Germany)~\cite{lago2021forecasting}.
	For each market, the target is the hourly day-ahead electricity price series, and two horizon-aligned exogenous series are provided as covariates (e.g., next-day load forecasts and wind-power related signals, depending on the market).
	EPF exhibits strong seasonality (daily/weekly cycles), frequent price spikes, and regime shifts driven by demand--supply balance and renewable variability, making it a challenging yet practical testbed for evaluating how future-known covariates improve short-horizon operational forecasts.
	
	\paragraph{\textbf{Energy} (Chilean national grid generation).}
	We adopt the Energy dataset released by the Chilean national grid operator, which records hourly generation from multiple sources~\cite{qiu2025dag}.
	Thermoelectric generation is treated as the target series, while the remaining sources (battery storage, wind, hydroelectric, solar, and geothermal) are used as covariates.
	This dataset is covariate-rich and characterized by strong cross-source coupling (e.g., renewables and storage affecting thermal dispatch), as well as non-stationary patterns across seasons. It provides a complementary benchmark to EPF by emphasizing multi-source interactions and covariate dependencies beyond price dynamics.
	
	\paragraph{\textbf{PV power} (industrial).}
	The photovoltaic (PV) dataset is collected from a real PV plant and targets multi-inverter active power forecasting.
	We include 23 inverter channels as targets and 8 on-site sensor covariates (e.g., irradiance- and meteorological-related measurements).
	The data are sampled at a 5-minute resolution and feature pronounced diurnal periodicity (near-zero generation at night) and sharp intra-day ramps under changing irradiance conditions.
	In addition to historical on-site measurements, reliable meteorological forecasts are incorporated as future-known covariates for long-term prediction, reflecting the operational setting where forecast quality depends critically on horizon-aligned weather inputs.
	
	\paragraph{\textbf{Day-ahead electricity price} (industrial).}
	This dataset is collected from an operational day-ahead price forecasting pipeline.
	The target is the day-ahead electricity price at a 15-minute resolution.
	We include 22 future-known covariates available prior to market clearing, consisting of (i) next-day dispatch disclosures that represent planned supply and demand conditions and (ii) meteorological forecasts that capture renewable-driven uncertainty.
	Compared with EPF, this dataset more explicitly reflects real-world operational constraints and heterogeneous exogenous sources, and it evaluates whether combining dispatch plans with weather-driven signals improves robustness across different day types (e.g., weekdays vs.\ weekends) and better captures intraday peak--valley dynamics.

	\subsection{Covariates: Availability and Screening}\label{sec:A2}
	\paragraph{\textbf{Covariate definition and availability.}}
	We divide exogenous variables into two groups: (i) past covariates $\bm{X}^{past}\in\mathbb{R}^{M_p\times T}$ observed over the same historical window as the target, and (ii) future-known covariates $\bm{Y}^{future}\in\mathbb{R}^{M_f\times F}$ that are available over the forecasting horizon.
	All covariates are temporally aligned to the target timeline and resampled to a consistent sampling frequency when needed.
	For horizon-aligned future-known covariates (e.g., weather forecasts and pre-announced dispatch plans), we strictly enforce an availability constraint: at inference time, only information that would be known when the forecast is issued is provided to the model, preventing look-ahead leakage and matching real deployment.
	
	This constraint is applied consistently in our industrial datasets. In long-term PV power forecasting, on-site environmental sensors (irradiance, temperature, humidity, wind) provide historical measurements only and are therefore treated as past covariates, while future inputs are limited to ex-ante meteorological forecasts.
	In day-ahead electricity price forecasting, future-known covariates are restricted to variables published prior to market clearing (e.g., next-day dispatch disclosures) together with meteorological forecasts, and no ex-post realized signals are used.

	\paragraph{\textbf{Covariate screening and driver analysis.}}
	To identify informative exogenous drivers and improve robustness, we perform lightweight screening before model training.
	
	\emph{PV power forecasting.}
	Since solar irradiance is the dominant physical driver of PV generation, we prioritize screening to retain the most informative on-site measurements and forecast signals.
	We use (i) daytime Pearson correlation to measure synchronous relationships with PV active power (restricting to daytime periods avoids trivial correlations induced by nighttime zeros), and
	(ii) Granger causality tests to assess whether the history of a covariate provides predictive information beyond an autoregressive baseline.
	Across the site data, irradiance-related variables consistently emerge as the strongest drivers, while temperature- and humidity-related variables contribute complementary information, especially under changing weather conditions that affect panel efficiency and atmospheric attenuation.
	
	\emph{Day-ahead electricity price forecasting.}
	We apply (i) Pearson correlation to quantify synchronous linear associations between candidate covariates and day-ahead prices, and
	(ii) Lasso-based importance by fitting a lag-augmented linear model and ranking covariates via the summed absolute coefficients over lags.
	Both analyses consistently highlight that the directly-dispatched load and the bidding space signal from the disclosed components are primary drivers, while irradiance-related features provide additional explanatory power by capturing renewable variability and uncertainty.

	\subsection{Implementation Details}\label{sec:A3}
	
	\paragraph{\textbf{Environment and optimization.}}
	We implement LightGTS-Cov in PyTorch and run all experiments on a single NVIDIA A100 GPU (80 GB).
	Unless otherwise specified, we optimize using Adam with an initial learning rate of $2\times10^{-4}$ and apply a StepLR learning-rate schedule during training.
	All models are initialized from a pretrained LightGTS backbone (LightGTS-tiny, $\sim$1M parameters).
	We report Mean Squared Error (MSE) and Mean Absolute Error (MAE) for evaluation.
	
	\paragraph{\textbf{Preprocessing and normalization.}}
	We apply consistent preprocessing across datasets.
	Missing timestamps are handled via time-index alignment; when necessary, short gaps in covariates are imputed using forward-fill.
	All continuous variables are normalized using statistics computed on the training split only, and the same transformation is applied to validation and test splits to prevent leakage.
	For multi-target settings (e.g., PV inverters), normalization is performed per target channel.
	
	\paragraph{\textbf{Evaluation protocol.}}
	For EPF and Energy, we follow the standard short-term protocol with lookback length $168$ and forecast horizon $24$.
	Metrics (MSE/MAE) are computed on the test split and averaged over all evaluation windows.
	For industrial datasets, forecasting horizons follow the operational requirements described in Sections~4.3--4.4, and we keep the covariate set identical across compared methods whenever applicable to ensure a fair comparison.

\end{document}